\documentclass[10pt,twocolumn,letterpaper]{article}

\usepackage{iccv}
\usepackage{times}
\usepackage{epsfig}
\usepackage{graphicx}
\usepackage{amsmath}
\usepackage{amssymb}
\usepackage{multirow}
\usepackage{pifont}
\newcommand{\cmark}{\ding{51}}
\newcommand{\xmark}{\ding{55}}
\usepackage{arydshln}
\usepackage{enumitem}

\usepackage[breaklinks=true,bookmarks=false]{hyperref}

\iccvfinalcopy 

\def\iccvPaperID{7028} 

\begin{document}

\title{GHuNeRF: Generalizable Human NeRF from a Monocular Video}

\author{Chen Li \quad \quad Jiahao Lin \quad \quad Gim Hee Lee\\
Department of Computer Science, National University of Singapore\\
{\tt\small lichen@u.nus.edu \quad \quad jiahao.lin@u.nus.edu \quad \quad gimhee.lee@comp.nus.edu.sg}
}

\maketitle
\ificcvfinal\thispagestyle{empty}\fi

\begin{abstract}

In this paper, we tackle the challenging task of learning a generalizable human NeRF model from a monocular video. 
Although existing generalizable human NeRFs have achieved impressive results, they require muti-view images or videos which might not be always available. On the other hand, some works on free-viewpoint rendering of human from monocular videos cannot be generalized to unseen identities. In view of these limitations, we propose GHuNeRF
to learn a generalizable human NeRF model from a monocular video of the human performer. 
We first introduce a visibility-aware aggregation
scheme to compute vertex-wise features, which is used to construct a 3D feature volume.  
The feature volume can only represent the overall geometry of the human performer with insufficient accuracy due to the limited resolution. To solve this, 
 we further enhance the volume feature with temporally aligned point-wise features using an attention mechanism. Finally, the enhanced feature is used for predicting density and color for each sampled point. A surface-guided sampling strategy is also adopted to improve the efficiency for both training and inference. We validate our approach on the widely-used ZJU-MoCap dataset, where we achieve comparable performance with existing multi-view video based approaches. We also test on the monocular People-Snapshot dataset and achieve better performance than existing works when only monocular video is used.  Our code is available at the project website\footnote{\url{https://github.com/chaneyddtt/GHuNeRF}}.
\end{abstract}

\section{Introduction}
Free-viewpoint synthesis of human performers has wide applications such as virtual reality, movie production, gaming, \etc. 
Traditional methods generally rely on images captured from dense camera views \cite{debevec2000acquiring, guo2019relightables} or accurate depth information \cite{collet2015high, dou2016fusion4d, su2020robustfusion}, which 
are tedious and expensive to obtain. Recently, neural radiance fields (NeRF) \cite{mildenhall2021nerf} is proposed to represent a scene as a continuous 5D function, and has achieved high-fidelity rendering results. However, 
vanilla NeRF 
inherits the limitation of the requirement for dense camera views, and furthermore requires computationally costly and non-generalizable per-scene optimization.

 \begin{figure}[t!]
\begin{center}
\includegraphics[width=0.95\linewidth]{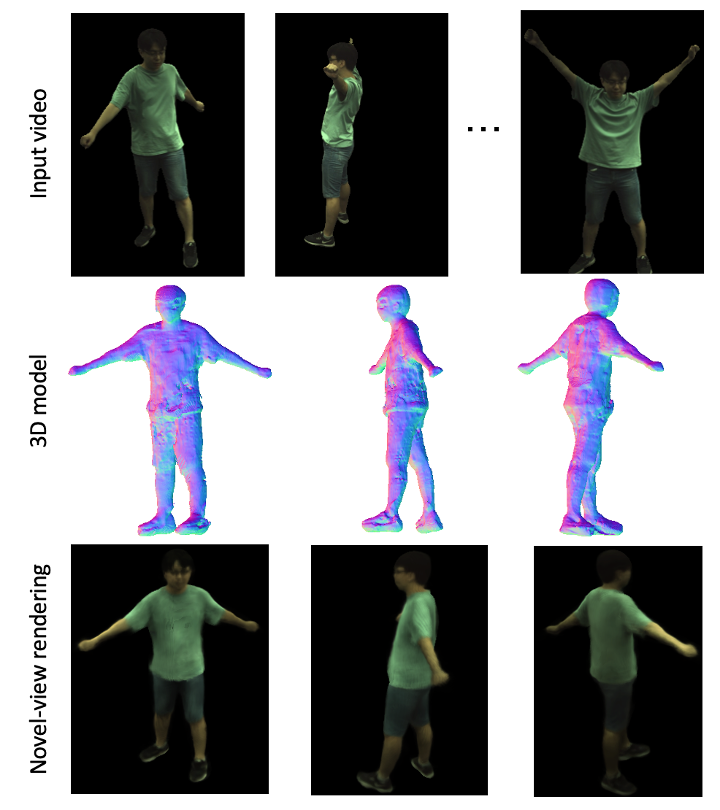}
\end{center}
\vspace{-3mm}
   \caption{An illustration of our task. We aim to construct a 3D human NeRF model that can be used to render free-viewpoint images from a monocular video of a performer.}
\label{fig:teaser}
\end{figure}

Recent human NeRFs \cite{peng2021neural, weng2022humannerf, jiang2022neuman} tackle the first limitation by aggregating information from videos to compensate for the lack of dense-view images. The aggregation of information is achieved by attaching a set of latent codes to the SMPL vertices \cite{loper2015smpl} or by constructing a canonical space. Impressive results have been achieved by these methods with a very sparse-view setting or even monocular videos. However, these methods still need per-subject training and cannot 
generalize to unseen subjects during test. 
To mitigate the generalization issue, generalizable human NeRFs \cite{zhao2022humannerf, kwon2021neural, chen2022geometry, mihajlovic2022keypointnerf} have been proposed to avoid per-subject training. 
The core idea is to take 
pixel-aligned features as input instead of the position information in the original NeRF formulation. Although sparse-view synthesis has been achieved by leveraging 
prior knowledge from pre-learned SMPL model, existing generalizable human NeRFs still require multi-view images or videos for both training and test. Unfortunately, this multi-view set-up is not always available in practice. 
In this paper, we aim to learn a 
generalizable human NeRF from monocular videos 
to overcome both 
the generalization issue and multi-view limitation. An illustration of our task is shown in Figure \ref{fig:teaser}.

One challenge of learning human NeRF from monocular videos is 
the modeling of large human motions. To this end, we make use of the parametric SMPL model to construct a feature volume. The feature representation for each SMPL vertex can be obtained by projecting it into 2D image space. Given that a body part is not always visible across the whole video due to occlusion, we propose a visibility-aware feature aggregation to extract useful information from each observed frame.  
This vertex-wise feature is then diffused to the whole feature volume by applying SparseConvNet \cite{graham20183d}. However, the feature volume 
can only represent the overall human geometry 
with insufficient accuracy because of the sparsity of the SMPL vertices and the limited volume resolution. To overcome this, we further enhance the volume feature at each location with a point-wise image feature. 
The point-wise feature is 
easy to obtain in the \textit{multi-view setting} by projecting a 3D point in the target space to the observed views. In contrast, we cannot directly use the 3D-to-2D projection in our \textit{monocular video setting} since the human is moving across the video and the corresponding 3D point in the observed space is unknown. 
We solve this issue by learning a transformation mapping from the target frame 
to the observed frames. The transformation is computed based on Linear Blend Skinning \cite{lewis2000pose}, where the blend weights are initialized with the SMPL model and further refined with a refinement network. Finally, we fuse the volume feature and the point-wise feature with an attention mechanism. 

We further adopt a surface-guided sampling strategy to improve the efficiency for both training and inference. Instead of randomly sampling points along a ray between a near and a far point as done in the 
vanilla NeRF, we sample points around the surface region to save memory and computation. Moreover, this also helps to regularize the 3D geometry implicitly since we are assuming that far away regions are empty space. 
We demonstrate the effectiveness of our approach on the widely used ZJU-MoCap dataset \cite{peng2021neural}, where we achieve comparable performance with existing multi-view video based approaches.  We also test on the monocular People-Snapshot dataset \cite{alldieck2018video} and achieve better performance. Our main contributions 
are as follows:
\begin{itemize} [leftmargin=0.35cm] 
    \item To the best of our knowledge, we are the first to tackle the task of learning a generalizable human NeRF model from monocular videos.
    \item We introduce GHuNeRF which consists of a visibility-aware volume feature aggregation and temporal aligned feature enhancement to aggregate information across video frames for free-viewpoint image synthesis. 

    \item We achieve state-of-the-art performance when only monocular video is available, and comparable performance with existing approaches that use multi-view videos.
\end{itemize}

\section{Related Work}
\paragraph{3D human reconstruction.}
3D human reconstruction from images is an extensively studied problem in computer vision. Early works \cite{bogo2016keep, lassner2017unite, kolotouros2019convolutional, kolotouros2019learning, pavlakos2019expressive} reconstruct 3D human body by fitting a parametric 3D human models \cite{loper2015smpl, joo2018total} to the input data. The parametric model makes it possible to recover both shape and pose information with only 2D supervision such as keypoints, silhouette and body segmentations. However, The parametric models are learned from minimally clothes body data, and hence cannot generalize well to clothed people. To solve this, implicit representation based approaches \cite{saito2019pifu, saito2020pifuhd, huang2020arch, he2021arch++} are proposed which can represent various topologies. PIFu and its variants \cite{saito2019pifu, saito2020pifuhd} adopt the occupancy field and reconstruct detailed surface geometry from even one view. ARCH and ARCH++ \cite{huang2020arch, he2021arch++} combine the parametric model with the
implicit field to estimate animatable 3D human avatars. Despite the high-fidelity reconstruction, these methods require the 3D ground truth for training which are expensive to collect. In comparison, we aim to achieve 3D human reconstruction and rendering from monocular videos. 

\vspace{-3mm}
\paragraph{Neural scene representation and rendering.} 
Recently, various neural representations \cite{lombardi2019neural, mildenhall2021nerf, thies2019deferred, zhou2018stereo, wu2020multi} have been introduced for novel view synthesis and geometric reconstruction. In particular, Neural radiance field (NeRF) \cite{mildenhall2021nerf} represents a scene as a continuous
5D function and has achieved high-fidelity rendering results.
Although effective, one main limitation of the vanilla NeRF is the requirement for expensive per-scene optimization. To mitigate this, subsequent works \cite{chen2021mvsnerf, johari2022geonerf, wang2021ibrnet, yu2021pixelnerf} learn generalizable NeRFs across different scenes by taking the pixel-aligned features as conditional information. Specifically, MVSNeRF \cite{chen2021mvsnerf} leverages plane-swept cost volumes to construct a neural encoding volume with per-voxel neural feature. An MLP is then adopted to regress volume density and color by using features interpolated from the encoding volume. IBRNet \cite{wang2021ibrnet} introduces a ray transformer to aggregate information from
nearby source views along a given ray. Inspired by these works, we also design a generalizable NeRF to model human body across different identities. This is more challenging comparing to existing generaliable NeRF \cite{chen2021mvsnerf, johari2022geonerf, wang2021ibrnet, yu2021pixelnerf} for static scenes since we need to model the large motion of human body simultaneously.  

\vspace{-3mm}
\paragraph{Neural radiance fields for human.}
To model the motion of human body, existing human NeRFs \cite{su2021nerf, xu2021h, peng2021neural, liu2021neural, jiang2022neuman, peng2021animatable, weng2022humannerf} rely on human-prior information, such as a skeleton or a parametric model. 
 NeuralBody \cite{peng2021neural} attaches a set of latent codes to the vertices of the SMPL model, which is able to aggregate information across different video frames. The per-vertex latent code is then diffused to generate a continuous latent code volume, which is used to regress the density and color for volume rendering. The other line of works \cite{xu2021h, liu2021neural,jiang2022neuman, peng2021animatable, weng2022humannerf} map all observations to a shared canonical space to model the large deformation of human bodies. HumanNeRF \cite{weng2022humannerf} adopts the inverse linear-blend skinning, which is combined with a non-rigid deformation, to learn a motion field mapping from observation to canonical space.  Impressive results have been achieved by HumanNeRFs even when using the monocular video as the inputs. However, these methods are designed for the person-specific setup, where a model needs to be trained for each identity. To solve this problem, recent works \cite{zhao2022humannerf, kwon2021neural, chen2022geometry} consider generalizable human rendering by taking the pixel-aligned features as conditions to avoid the memorization of human-specific density and color. 
 GPNeRF \cite{chen2022geometry} proposes a geometry-guided progressive rendering mechanism, which leverages the geometry volume and the predicted density to reduce the number of sampling points.
 The most related work is NHP \cite{kwon2021neural}, which designs a temporal transformer to aggregate tracked visual features based on the skeletal body motion. The temporally-fused features are then merged with multi-view pixel-aligned features for novel view synthesis. Although NHP achieves cross-identity and cross-dataset generalization, it requires multi-view videos which might not be always available in practice. In this paper, we aim to learn a generalizable human NeRF model with only monocular video as the inputs. 

 \begin{figure*}[t]
\begin{center}
\includegraphics[width=0.99\linewidth]{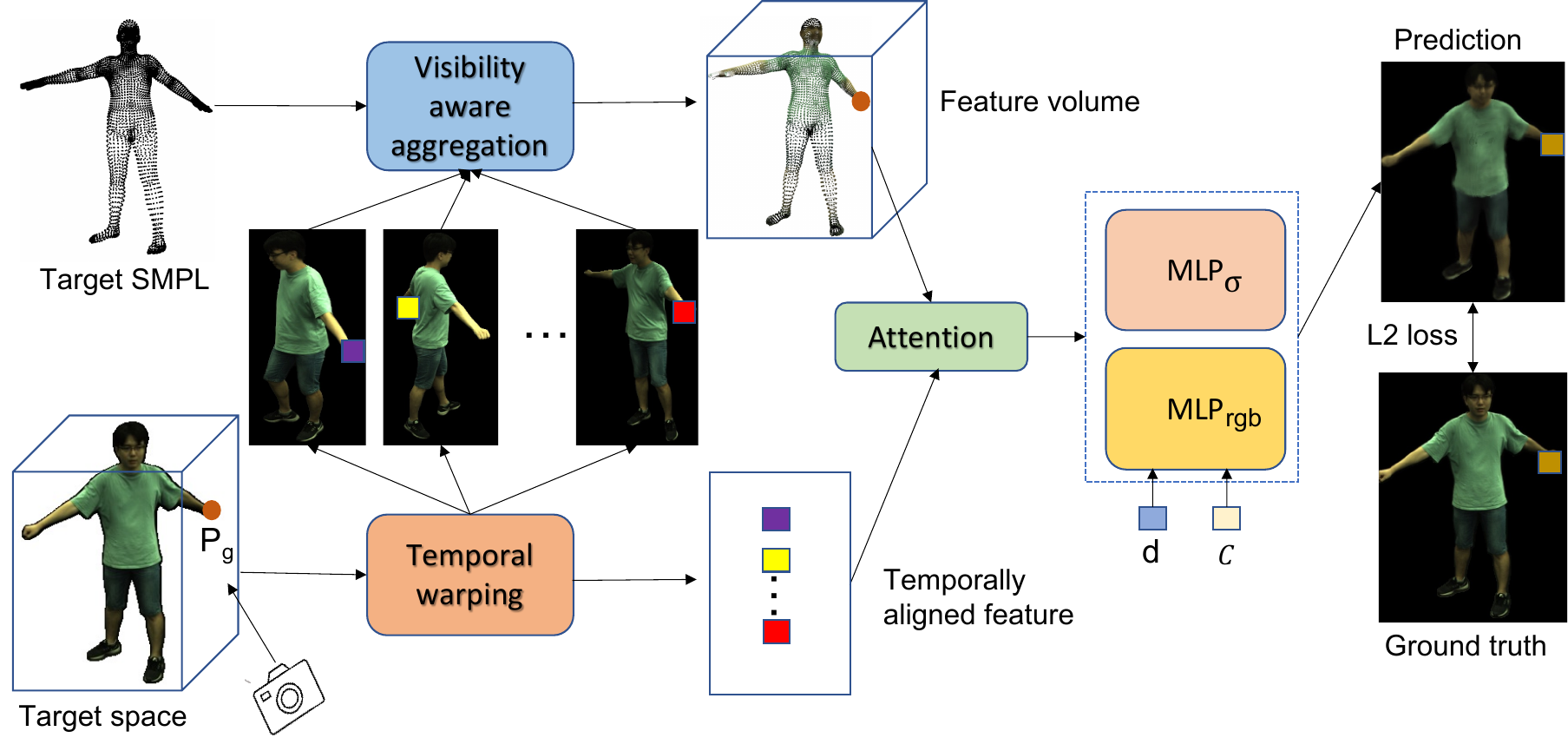}
\end{center}
\vspace{-3mm}
   \caption{The overall pipeline of our GHuNeRF framework. We first compute the feature representation for each vertex of the target SMPL using the visibility-aware feature aggregation. A feature volume is then constructed based by diffusing the vertex-wise feature to nearby 3D space. The volume feature is further enhanced with the temporally aligned point-wise feature using the attention mechanism. Finally, the enhanced feature is used for predicting density and color for rendering.}
\label{fig:short}
\end{figure*}

\section{Our Method: GHuNeRF}
We propose GHuNeRF to construct a 3D human model that can be used to render free-viewpoint images from a monocular video of a performer. 
The overall pipeline of our GHuNeRF is shown in Figure~\ref{fig:short}.
To handle the large motions of human body, we leverage the SMPL model to aggregate vertex-wise feature from input video frames. Specifically, we introduce a \textbf{visibility-aware aggregation} scheme 
since a vertex can be
observed from only some of the frames due to occlusion. We then construct a feature volume from the SMPL vertices features using SparseConvNet.  
 The volume information can only represent the overall geometry, but 
is not sufficiently accurate due to the sparsity of the SMPL vertices and the limited resolution of the volume. To solve this, we further enhance the volume features with \textbf{temporally aligned} point-wise features based on an attention mechanism. This enhanced feature is used to predict the color and density for novel-view rendering. 
 Moreover, we also introduce a \textbf{surface-guided point sampling strategy}, where only points around the surface region are sampled to improve the efficiency. 

\subsection{Visibility-aware Volume Feature Aggregation}
We denote the observed video as $I_{1:T} = \{I_1, I_2, ..., I_T\}$, where $T$ is the number of frames. The objective is to reconstruct a 3D human model that can be rendered at any time step $t$ at any camera view $c$, which we denote as target frame $I_g$. To synthesize the target frame, we construct a feature volume in the space of the target frame, \ie target space, based on the target SMPL parameters. Specifically, for each vertex on the target SMPL, we obtain the feature representation by aggregating information from the observed frames. Visibility information is incorporated during the aggregation since a vertex can be visible to only some of the observed frames due to occlusion. The aggregated feature for each vertex is computed as:
\begin{align}
    F(v_g) = \frac{\sum_{i=1}^T b_i \times F(v_i)}{\sum_{i=1}^T b_i }, 
    \label{vertex_feature}
\end{align}
where $v_g$ denotes a vertex on the target SMPL and $\{v_1, v_2, ..., v_T\}$ denote the corresponding SMPL vertex of the observed frames, $b_i$ represents the visibility of the vertex $v_i$. The feature representations of the vertices in the observed frames are computed by projecting them into the 2D image space.
Different from previous NHP  \cite{kwon2021neural} and GPNeRF \cite{chen2022geometry} which the use attention mechanism \cite{vaswani2017attention} to predict scores for feature aggregation, our visibility information for each vertex is obtained from a rasterization process. 
By considering the visibility, we are able to collect useful information across different times steps 
which thus compensates for the lack of multi-view information.

The vertex-wise feature representation is sparse and does not fulfill the requirement for volume rendering, where the density and color are queried in a continuous 3D space. We further use the SparseConvNet \cite{graham20183d} to diffuse the vertex-wise feature to the nearby 3D space following \cite{peng2021neural}. Specifically, a 3D bounding box of the human in the target frame is computed based on the corresponding SMPL parameters, and the box is gridded into a tessellation of smaller voxels with voxel size of $5\text{mm} \times 5\text{mm} \times 5\text{mm}$. The feature representation of any non-empty voxel is computed as the mean of features of SMPL vertices within this voxel.

\subsection{Temporally Aligned Feature Enhancement}
The feature volume constructed based on SMPL vertices can represent the overall structure of a human, but is 
not sufficiently accurate due to the sparsity of the SMPL vertices and the limited volume resolution.  To overcome this problem, we propose to enhance the volume features with temporally aligned point-wise features. In the multi-view based approaches \cite{zhao2022humannerf, kwon2021neural, chen2022geometry}, this point-wise feature can be easily obtained by projecting any sampled point in the target space into observed views with the known camera parameters. However, in our monocular video setting, we 
cannot directly project any 3D sampled point in the target space to the observed video frames since we do not know the corresponding 3D point in the observation space. To solve this, 
 we define a transformation field mapping 3D sampled points from the target space to the observation space. 

It is hard to directly model the transformation with deep networks given the large motion of human body. We instead formulate the transformation based on Linear Blend Skinning. The Linear Blend Skinning is commonly used in character animation, where the transformation of each vertex is affected by all body parts. Specifically, a vertex $v$ on a template mesh can be posed
via the Linear Blend Skinning as:
\begin{align}
    v' = (\sum_{j=1}^N w_j \mathcal{T}_j) v.
\end{align}
$\mathcal{T}_j$ is the transformation for the $j^\text{th}$ body part from a template pose and $w_j$ represents the corresponding blend weight. $N$ denotes the total number of body parts. In our case, the transformation $\mathcal{T}$ for each body part can be computed from the pose parameter $\Theta$ and the joint locations $J$ in the template pose based on SMPL.  
Given the target pose $\Theta_g$ and the observed pose $\Theta_o$, the transformation from the target space to the observation space can then be expressed as:
\begin{align}
    p_o = (\sum_{i=1}^N w_g \mathcal{T}_o \mathcal{T}_g^{-1}) p_g,
    \label{g2o-warpping}
\end{align}
where $w_g$ represents the blend skinning weights in target space. 
We 
follow previous works \cite{peng2021animatable, huang2020arch, bhatnagar2020loopreg} to model the blending weights by leveraging prior knowledge from the SMPL model. Specifically, for any 3D sampled point in the target space, we assign the initial blending weight as the same value as the nearest vertex on the SMPL surface. These initial blending weights are generally inaccurate, especially for 3D points that are far from the surface. 
To improve the accuracy, we further use an MLP network for refinement, \ie: 
\begin{align}
    w_g = w_s + \text{MLP}_w (w_s, \Theta, d),
\end{align}
where $w_s$ denotes the initial blending weight. $\text{MLP}_w$ represents the refinement network, which takes 
the initial weight $w_s$, the pose parameters $\Theta$ and the distance to the nearest surface point $d$ as inputs. Intuitively, the blend skinning weight of a 3D point in the target space depends on both the target body pose and the distance from this point to the body surface. We apply softmax to the output of the refinement network such that the weights for different body parts sum up to one. 

For any 3D sampled point $p_g$ in the target space, we can warp it to the observation spaces of all input frames using Eqn. \eqref{g2o-warpping}, denoted as $\{p_0^o, p_1^o, ..., p_T^o\}$. The corresponding point-wise feature can then be extracted as $\{F (p_1^o), F (p_2^o), ..., F (p_T^o)\}$ via a 3D-to-2D projection. We use these point-wise features to enhance the volume features. Specifically, the volume feature of a 3D point $p_g$ can be retrieved from the volume using bilinear interpolation, which we denote as $F_v (p_g)$.  This volume-based feature $F_v (p_g)$ is enhanced with the temporally aligned point-wise features with the attention mechanism \cite{vaswani2017attention}:
\begin{align}
    F_e (p_g) = \text{Attention} ( & Q = F_v (p_g), \\ \notag
   & K = \{F (p_t^o)\}_{t=1}^T, \\ \notag
    & V = \{F ((p_t^o)\}_{t=1}^T).
\label{feature_enhancement}
\end{align}
Intuitively, the attention mechanism helps to incorporate relevant information from the input frames and ignore the irrelevant ones.  

Finally, the enhanced feature is used to predict the density and color of each sampled point:
\begin{align}
    & \sigma(p_g) = \text{MLP}_\sigma (F_e (p_g)), \\ \notag
    & \text{c} (p_g) = \text{MLP}_\text{rgb} (F_e (p_g), \gamma_\mathbf{d} (\mathbf{d}), \mathcal{C}),
\label{final_prediction}
\end{align}
where $\text{MLP}_\sigma$ and $\text{MLP}_\text{rgb}$ are the density and color prediction networks, respectively, and $\gamma_\text{d}$ represents the positional encoding for view direction $\text{d}$. $\mathcal{C}$ is a per-camera latent code to encode the camera-specific elements.

\subsection{Surface-guided Points Sampling}
We further adopt a surface-guided sampling strategy to replace the random sampling used in the original NeRF. The motivation is twofold. First, the random sampling 
inevitably leads to numerous 
sampled points in the empty space 
and thus slowing down the convergence of the network. In contrast, 
our surface guided sampling strategy only samples points near the SMPL surface region, which significantly reduces the number of unnecessary points in the empty space and thus improving the efficiency. Second, 
our surface-guided sampling is able to regularize the 3D geometry implicitly since we are assuming that points far away from the surface corresponding to empty space. 
Computing the distance between a sampled point to the SMPL surface during training is expensive. 
To reduce computations, we voxelize the 3D space and pre-compute the distance between each voxel and the SMPL surface. The distance between any sampled 3D point and surface is computed with bilinear interpolation during training.  

\subsection{Volume Rendering}
We use the volume rendering to render the RGB values for each pixel at time step t in the target view:
\begin{align}
    \Tilde{C}_t(\mathbf{r}) &= \sum_{k=1}^{N_k} T_k(1-\text{exp}(-\sigma_k \delta_k))\text{c}_k, \\
    &\text{where} \quad T_k = \text{exp}(-\sum_{j=1}^{k-1}\sigma_j\delta_j).
\end{align}
$N_k$ denotes the number of sampling points along each ray and $\delta_k$ is the distance between adjacent sampled points. 

The objective function is the squared error between the rendered color $\Tilde{C}_t(\text{r})$ and the ground truth color $ C_t(\mathbf{r})$:
\begin{align}
    \mathcal{L} = \sum_{\mathbf{r} \in \mathcal{R}}\|\Tilde{C}_t(\mathbf{r}) - C_t(\mathbf{r})\|_2^2.
\end{align}
During training, our network can be supervised with both multi-view or monocular videos. In the multi-view training (MVT) setting, we aim to synthesis an image at any time step from a different view. In the monocular training (MoT) setting, we aim to synthesis an image at any time step in the same view.  It should be noted that only a monocular video of a human performer is used as inputs during inference in both settings. 

\section{Experiments}
\subsection{Implementation details}
\paragraph{Networks details.} We use the ResNet18 \cite{he2016deep} following \cite{kwon2021neural} to extract image features, which are used to compute both vertex-wise feature in Eqn. (1) and point-wise feature in Eqn. (5). The blend weights refinement MLP consists of eight layers with the channel size of 256. The SparseConvNet consists of four blocks of convolution and downsampling layers 
with 2×, 4×, 8×, 16× downsampled sizes. The attention is performed twice for the density feature and color feature respectively. The per-camera latent code has a dimension of 128 and is optimized together with the network.
Note that each video in both ZJU-MoCap and People-Snapshot datasets contains hundreds of frames, and we select 15 frames from the whole video as the inputs for both memory and computation efficiency. More details are provided in the 
supplementary material.  

\paragraph{Training details.} We adopt the Adam optimizer \cite{kingma2014adam} for training with a learning rate of $1e^{-4}$ and a batch size of one. Both training and inference are conducted with an image size of 512 × 512 for fair comparison with previous works. We train our network for 500 epochs with 500 iterations in each epoch on one RTX 3090Ti GPU.  The distance threshold for 
our surface-guided points sampling is set to 5 cm. 

\subsection{Datasets and Evaluation Metrics}
We evaluate our approach on the ZJU-MoCap dataset \cite{peng2021neural} and the People-Snapshot dataset \cite{alldieck2018video}. The ZJU-MoCap dataset consists of 9 dynamic human video with different appearance and poses. Each subject is captured using a multi-camera system with 21 synchronized cameras. We use 6 subjects for training and 3 subjects for evaluation. The People-Snapshot dataset is a monocular video dataset which captures performers that rotate while holding an A-pose. We randomly select 3 identities for testing and use the remaining for training. We evaluate our method under both MVT and MoT settings. We select 15 frames from the whole video sequence for each target image in the MVT setting, and always use the same 15 frames to synthesize all other frames in the same video in the MoT setting.
We adopt the peak signal-to-
noise ratio (PSNR) and structural similarity index (SSIM) as the evaluation metrics following \cite{peng2021neural, kwon2021neural}. 
We also provide qualitative results for 3D human reconstruction as there is no ground truth human geometry.

\subsection{Results: ZJU-MoCap Dataset}
We first evaluate our method on the most commonly used ZJU-MoCap dataset. We conduct experiments under both MVT and MoT settings, where the network is trained with multi-view and monocular data, respectively. Note that \textbf{we only require a monocular video} as the input during test under both settings.  We compare with previous human NeRF works including NT \cite{thies2019deferred}, NHR \cite{wu2020multi}, NV \cite{lombardi2019neural}, NB \cite{peng2021neural}, PVA \cite{raj2021pixel}, pixelNeRF \cite{yu2021pixelnerf}, KeypointNeRF \cite{mihajlovic2022keypointnerf}, GPNeRF \cite{chen2022geometry} and NHP \cite{kwon2021neural}. Among the comparison methods, NT, NHR, NV and NB are not generalizable which require per-scene optimization, PVA, KeypointNeRF, GPNeRF and NHP are generalizable, but use multi-view videos (NHP) or images (GPNeRF, PVA) as inputs for both training and test. 
The results for pixelNeRF are obtained under multi-view setting although the input for pixelNeRF can be both monocular or multi-view images.

We test our approach on both seen and unseen identities and the quantitative results are shown in Table \ref{results-on-seen-zju} and Table \ref{results-on-unseen-zju}, respectively. 
As can be seen from Table \ref{results-on-seen-zju}, our approach achieves comparable performance with NHP and GPNeRF, although we only use a monocular video as input during test while NHP and GPNeRF takes in multi-view inputs. Moreover, we also achieve comparable results with the optimization based approach Neuralbody \cite{peng2021neural}. For the results for the unseen identities in Table \ref{results-on-unseen-zju}, we outperform the optimization based approaches, which are optimized on the unseen subjects before inference. Comparable performance is also achieved compared with NHP and GPNeRF. Results in both tables verify the effectiveness of our approach. 

\begin{figure*}[ht!]
\begin{center}
\includegraphics[width=0.95\linewidth]{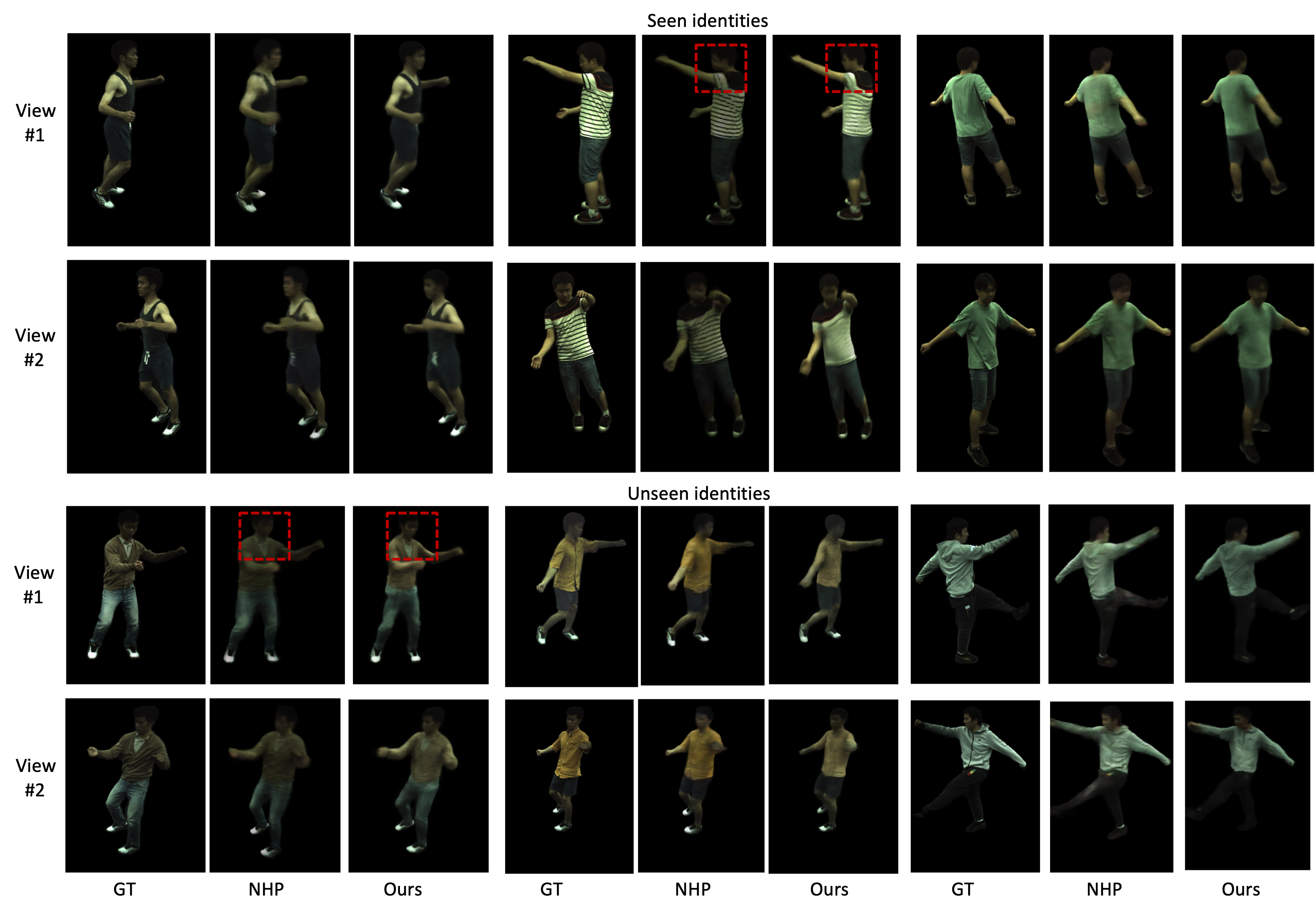}
\end{center}
\vspace{-2mm}
   \caption{Qualitative results for novel-view synthesis on the ZJU-MoCap dataset.}
\label{fig:qualitiative-results-zju}
\end{figure*}

\begin{figure}[ht!]
\begin{center}
\includegraphics[width=0.96\linewidth]{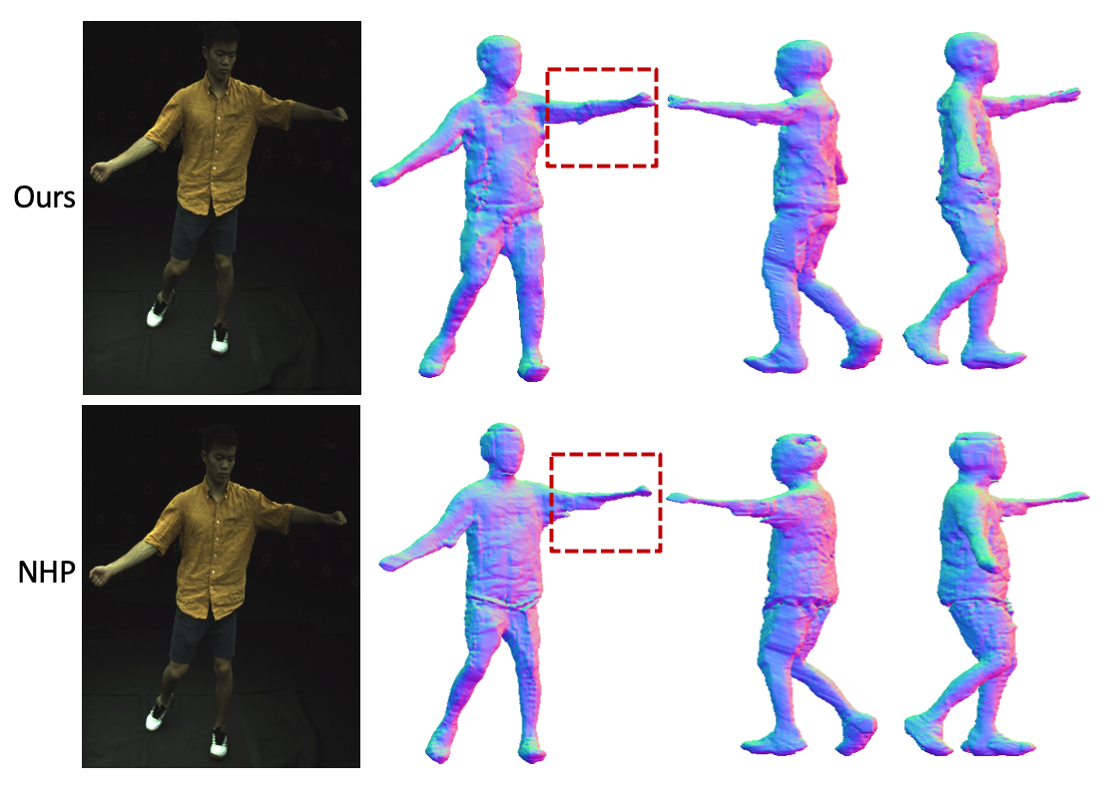}
\end{center}
\vspace{-2mm}
   \caption{Visualization of 3D human model on the ZJU-MoCap.}
\label{fig:qualitative-results-zju-3d}
\end{figure}

We show the qualitative results for both seen and unseen identities in Figure \ref{fig:qualitiative-results-zju}, where we compare with the most related NHP. We can see that our approach is able to synthesize as high-fidelity images 
as NHP although NHP uses multi-view videos as the input. Even better results are achieved in some cases, where our approach generates more details on the face (marked with red box). We also show the 3D human model predicted by our network in Figure \ref{fig:qualitative-results-zju-3d}. Our approach is able to predict realistic 3D human shapes in the case where NHP fails to predict the left arm of the performer. More qualitative results are provided in the supplementary material.

\begin{table}[ht!]
\centering
\small
\setlength{\tabcolsep}{4pt}
\begin{tabular*}{0.49\textwidth}{c|c|c c| c c}
\hline
  \multirow{2}{*} {Method} &  \multirow{2}{*} {Generalizable} & \multicolumn{2}{c|}{Monocular}  & \multicolumn{2}{c}{Seen subjects} \\
  & & Train & Test & PSNR  ($\uparrow$) & SSIM  ($\uparrow$) \\ \hline
NT \cite{thies2019deferred} & \xmark & \xmark & $\ast$ & 23.86  & 0.896  \\
NHR \cite{wu2020multi} & \xmark & \xmark & $\ast$ & 23.95  & 0.897 \\ 
NB\cite{peng2021neural} & \xmark & \xmark & $\ast$ & 28.51  & 0.947 \\ 
NHP \cite{kwon2021neural} & \cmark & \xmark & \xmark & 28.73 & 0.936 \\ 
GPNeRF \cite{chen2022geometry} & \cmark & \xmark & \xmark & 28.91  & 0.944  \\\hdashline
Ours-MVT & \cmark & \xmark & \cmark & 27.24 &  0.930 \\
Ours-MoT & \cmark & \cmark & \cmark & 27.32 & 0.936   \\\hline
\end{tabular*}
\caption{Quantitative results on the seen subjects of the ZJU-MoCap dataset. $\ast$ indicates method is trained on per-scene optimization and does not require image input during test.}
\label{results-on-seen-zju}
\end{table}

\begin{table}[ht!]
\centering
\small
\setlength{\tabcolsep}{3pt}
\begin{tabular*}{0.498\textwidth}{c|c|c c | c c}
\hline
 \multirow{2}{*} {Method} &  \multirow{2}{*} { Generalizable} & \multicolumn{2}{c|}{Monocular} & \multicolumn{2}{c}{Unseen subjects} \\
 
  & & Train & Test & PSNR ($\uparrow$) & SSIM ($\uparrow$) \\ \hline
NV \cite{lombardi2019neural} & \xmark &\xmark & $\ast$ & 20.84 & 0.827  \\ 
NT \cite{thies2019deferred} & \xmark & \xmark & $\ast$ & 21.92  & 0.873  \\
NHR \cite{wu2020multi} & \xmark & \xmark & $\ast$ & 22.03 & 0.875  \\ 
NB \cite{peng2021neural} & \xmark & \xmark & $\ast$ & 22.88 & 0.880 \\ 
PVA \cite{raj2021pixel} & \cmark & \xmark & \xmark & 23.15 & 0.866\\
pixelNeRF \cite{yu2021pixelnerf} & \cmark & \xmark & \xmark & 23.17 & 0.869 \\
KeypointNeRF \cite{mihajlovic2022keypointnerf} & \cmark & \xmark & \xmark & 25.03  & 89.69 \\ 
NHP \cite{kwon2021neural} & \cmark & \xmark & \xmark & 24.75 & 0.906 \\ 
GPNeRF \cite{chen2022geometry} & \cmark & \xmark & \xmark & 25.96 & 0.921   \\\hdashline
Ours-MVT & \cmark & \xmark & \cmark & 24.12 & 0.905  \\
Ours-MoT & \cmark & \cmark & \cmark & 24.55 & 0.911  \\\hline
\end{tabular*} 
\caption{Quantitative results on the unseen subjects of the ZJU-MoCap dataset. $\ast$ indicates method is trained on per-scene optimization and does not require image input during test. 
}
\label{results-on-unseen-zju}
\end{table}

\subsection{Results: People-Snapshot Dataset}
We also test our approach on the People-Snapshot dataset which only consists of monocular videos. Our approach is trained under the monocular setting (MoT) on this dataset since multi-view supervision is not available. We randomly select three subjects from the dataset as test subjects and the remaining are used for training. During inference, the same 15 frames are evenly selected from the whole video to synthesize the 
remaining video frames. We compare our approach with NHP \cite{kwon2021neural} \textbf{since NHP is also generalizable and can be trained with monocular videos}. Note that the results of NHP in Table~\ref{results-on-unseen-people-snapshot} are based on our implementation since the temporal aggregation component of NHP is not publicly available. As can be seen from Table~\ref{results-on-unseen-people-snapshot}, our approach outperforms NHP by a large margin when only monocular videos are available.

To evaluate the generalization capacity of our approach, we also show results for the cross-dataset generalization in Table~\ref{results-cross-dataset}. The results are obtained by directly applying our model trained on the People-Snapshot dataset to the ZJU-MoCap dataset. We can see that the cross-dataset generalization achieves similar PSNR score with our model trained on the ZJU-MoCap dataset, 23.20 vs 24.55.

\begin{table}[ht!]
\centering
\small
\setlength{\tabcolsep}{4pt}
\begin{tabular*}{0.49\textwidth}{c|c|cc|cc}
\hline
  \multirow{2}{*} {Method} &  \multirow{2}{*} { Generalizable} & \multicolumn{2}{c|}{Monocular}  & \multicolumn{2}{c}{Test subjects} \\
  & & Train & Test & PSNR ($\uparrow$) & SSIM ($\uparrow$) \\ \hline
NHP \cite{kwon2021neural} & \cmark & \cmark & \cmark & 26.22 & 0.903 \\ 
Ours-MoT & \cmark & \cmark & \cmark & 28.37 & 0.927  \\\hline
\end{tabular*}
\caption{Results on unseen subjects of People-Snapshot dataset.}
\label{results-on-unseen-people-snapshot}
\end{table}

 \begin{figure*}[h!]
\begin{center}
\includegraphics[width=0.95\linewidth]{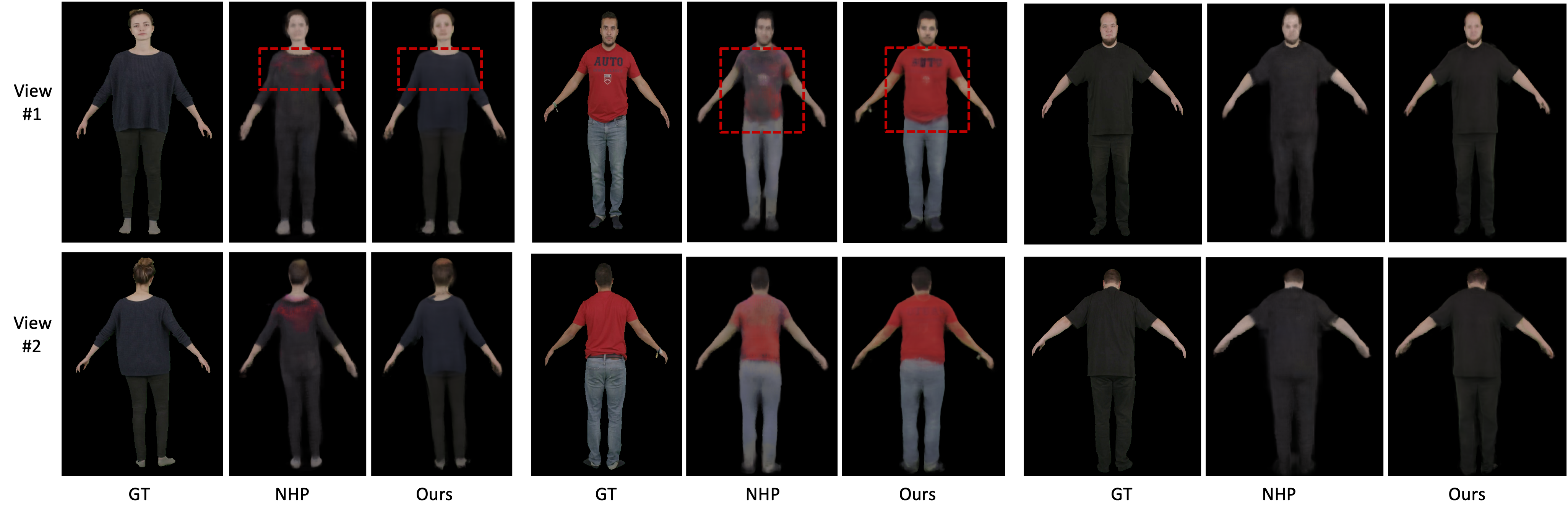}
\end{center} \vspace{-2mm}
   \caption{Qualitative results for novel view synthesis on the People-Snapshot dataset.}
\label{fig:qualitative-results-people-snapshot}
\end{figure*}

 \begin{figure}[h!]
\begin{center}
\includegraphics[width=0.96\linewidth]{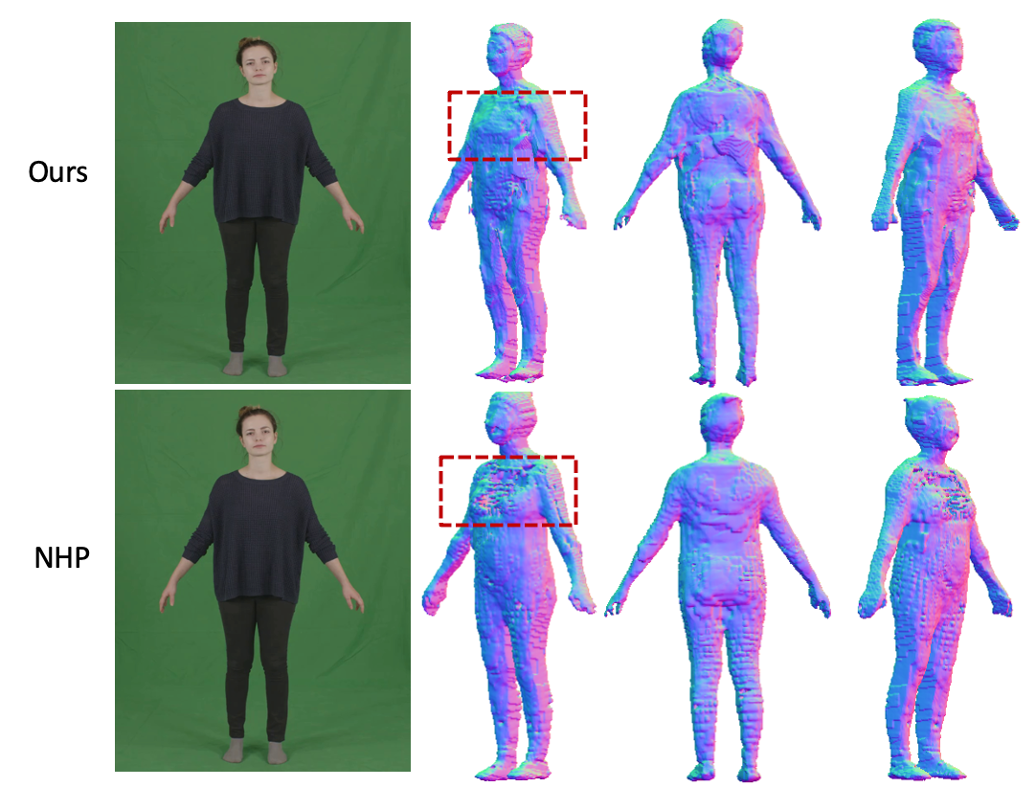}
\end{center}
\vspace{-2mm}
   \caption{Visualization of 3D human on the People-Snapshot.}

\label{fig:quiatative-results-people-snapshot-3d}
\end{figure}

We show the qualitative results for novel-view synthesis on the People-Snapshot dataset and compare with the most related NHP \cite{kwon2021neural} in Figure \ref{fig:qualitative-results-people-snapshot}. NHP fails to generate the correct color (marked in red box) for the target image when trained only with monocular videos. In comparison, our approach is able to synthesize realistic novel-view images across different identities. 
Qualitative results for the 3D human model predicted by our network are also provided in Figure \ref{fig:quiatative-results-people-snapshot-3d}. We can see that NHP struggles to estimate the 3D shapes, where 
holes on the chest of the performer can be observed. In contrast, our method estimates more realistic 3D human shapes with fewer artifacts.

\begin{table}[ht!]
\centering
\small
\setlength{\tabcolsep}{4pt}
\begin{tabular*}{0.49\textwidth}{c|c|c c|c c}
\hline
  \multirow{2}{*} {Method} &  \multirow{2}{*} { Generalizable} & \multicolumn{2}{c|}{Monocular} & \multicolumn{2}{c}{Test subjects} \\
 
 & & Train & Test & PSNR ($\uparrow$) & SSIM ($\uparrow$) \\ \hline
NHP \cite{kwon2021neural} & \cmark & \cmark & \cmark & 16.07 & 0.836 \\ 
Ours-MoT & \cmark & \cmark & \cmark & 23.20 & 0.889  \\\hline
\end{tabular*}
\vspace{-2mm}
\caption{Quantitative results for cross-dataset generalization.}
\label{results-cross-dataset}
\end{table}

\subsection{Ablation Studies}
We conduct ablation studies on the ZJU-MoCap by removing each component successively from the full model. The results are shown in Table \ref{ablation-studies}, where `VVF' denotes visibility-aware volume feature, `BWR' the blending weights refinement,  `TFE' the temporally aligned feature enhancement and `SGS' the surface-guided sampling. We first use the volume feature and remove the blending weights refinement, the temporally aligned feature enhancement and the surface-guided sampling successively. The performance drops when each component is removed, especially for the temporally aligned feature enhancement. To further verify the role of the visibility-aware volume feature, we then remove the volume feature, and instead use the temporally aligned feature for density and volume prediction. We directly take the mean value of the temporal features instead of using the attention operation in Eqn.(5). We can see that the performance drops when the volume feature is removed.

\begin{table}[h!]
\centering
\setlength{\tabcolsep}{8pt}
\begin{tabular*}{0.49\textwidth}{c c c c| c c}
\hline
   VVF & BWR &  TFE & SGS  &  PSNR ($\uparrow$) & SSIM ($\uparrow$)   \\ \hline
   \cmark  & \cmark & \cmark & \cmark & 24.12  & 0.905 \\ 
    \cmark  & \xmark & \cmark & \cmark & 24.06 &  0.905 \\
    \cmark  & \xmark & \xmark & \cmark & 22.94 & 0.896  \\
        \cmark  & \xmark & \xmark & \xmark & 22.56 & 0.890\\
      \xmark  & \cmark & \cmark & \cmark & 23.70 & 0.888 \\ \hline
\end{tabular*}
\caption{Ablation study of successive removal of each component.}
\label{ablation-studies}
\end{table}

\section{Limitations}
Our approach achieves impressive results 
on both novel-view synthesis and 3D human reconstruction, but there are still many challenges for the task of learning generalizable human NeRF from monocular videos. For example, the generalization capacity is still limited when the training and testing data are significantly different. Some failure cases are shown in our supplementary material.

\section{Conclusion}
We propose GHuNeRF to tackle the task of learning a generalizable human NeRF from monocular videos in this paper. 
We leverage the SMPL model to construct a feature volume where a visibility-aware feature aggregation scheme is introduced to integrate useful information from input frames. A volume feature enhancement is designed to enhance 
coarse volume feature with temporally-aligned point-wise feature. Moreover, a surface-guided sampling strategy is proposed to improve the efficiency for both training and inference. 
Extensive experiments have been conducted to validate the effectiveness of our approach.

\pagebreak


	\twocolumn[{%
		\renewcommand\twocolumn[1][]{#1}%
		\vskip .5in
		\begin{center}
			\textbf{\Large Supplementary Material for}\\
			\vspace*{4pt}
			\textbf{\Large GHuNeRF: Generalizable Human NeRF from a Monocular Video} \\
			\vspace*{10pt}
			{\large
				Chen Li \quad \quad Jiahao Lin \quad \quad Gim Hee Lee\\
			}
			\vskip .5em
			{\large Department of Computer Science, National University of Singapore\\}
			{\tt\small lichen@u.nus.edu \quad \quad jiahao.lin@u.nus.edu \quad \quad gimhee.lee@comp.nus.edu.sg}
			\vspace*{10pt}
		\end{center}
	}]

\setcounter{equation}{0}
\setcounter{section}{0}
\setcounter{page}{9}
\paragraph{Selection of video frames as the inputs.}
Each video in the ZJU-MoCap and People-Snapshot datasets contains hundreds of frames, which will cause memory issue if directly taking the whole video as input.  To solve this problem,
we select 15 frames from the whole video sequence based on two criteria: 
1) To select evenly from each video, denoted as criterion \#1. 
2) To select based on the SMPL vertices, denoted as criterion \#2. Specifically, we compute the distance between the SMPL vertices of the target frame and each video frame in the camera coordinate, and take the closest 15 frames as the inputs. Empirically, criterion \#2 performs slightly better than criterion \#1, as shown in Table \ref{results-different-criteria}. Intuitively, criterion \#2 
selects frames that have similar body direction and pose with the target frame, hence results in better performance.

\begin{table}[ht!]
\centering
\small
\setlength{\tabcolsep}{5pt}
\begin{tabular*}{0.49\textwidth}{c|c c | c c}
\hline
 \multirow{2}{*}{Selection criteria} & \multicolumn{2}{c}{Seen subjects} & \multicolumn{2}{c}{Unseen subjects}\\
 
 & PSNR ($\uparrow$) & SSIM ($\uparrow$) & PSNR ($\uparrow$) & SSIM ($\uparrow$) \\ \hline
Criterion \#1 & 27.19 & 0.930 & 23.71 & 0.902\\ 
Criterion \#2 & 27.24 & 0.930  & 24.12 & 0.905\\\hline
\end{tabular*}
\caption{Quantitative results based on different selection criteria.}
\label{results-different-criteria}
\end{table}

\begin{figure}[h!]
\begin{center}
\includegraphics[width=0.99\linewidth]{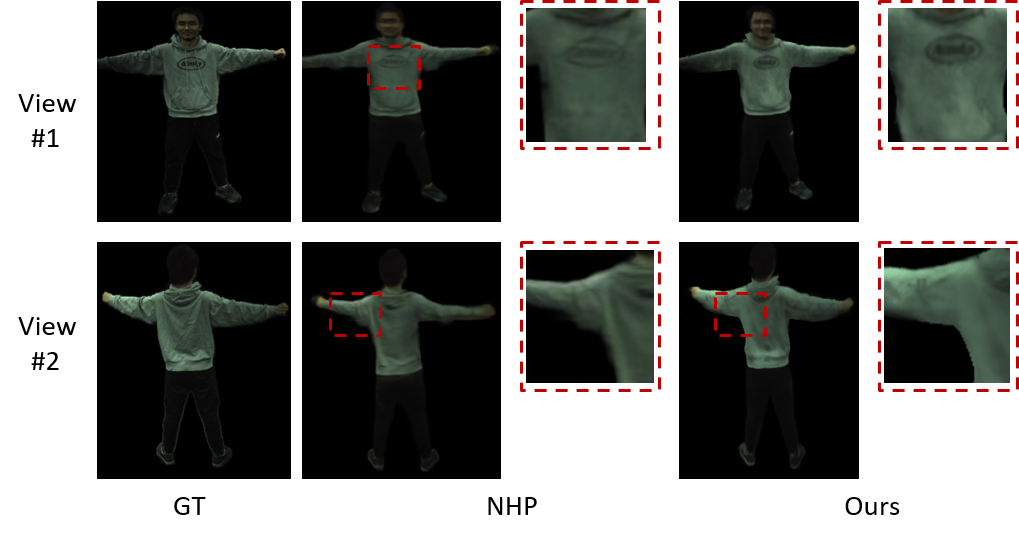}
\end{center}
   \caption{Qualitative comparison for unseen identities.}
\label{fig:qualitative-results-unseen-supp}
\end{figure}

\begin{figure*}[h!]
\begin{center}
\includegraphics[width=0.99\linewidth]{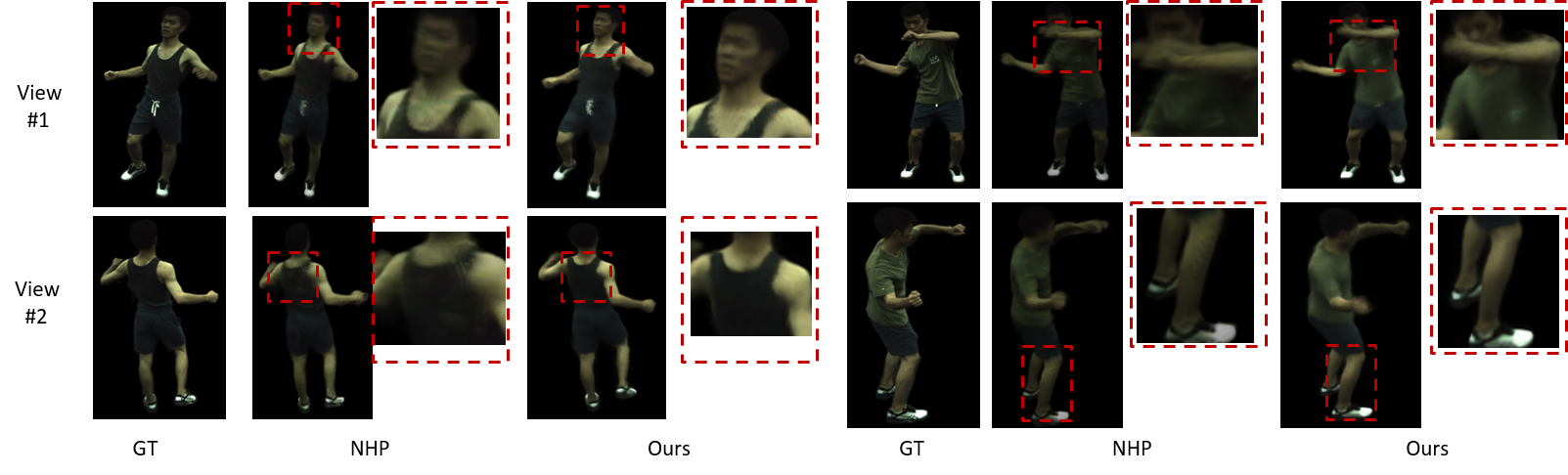}
\end{center} 
   \caption{Qualitative comparison for seen identities.}
\label{fig:qualitative-results-seen-supp}
\end{figure*}

\begin{figure*}[h!]
\begin{center}
\includegraphics[width=0.99\linewidth]{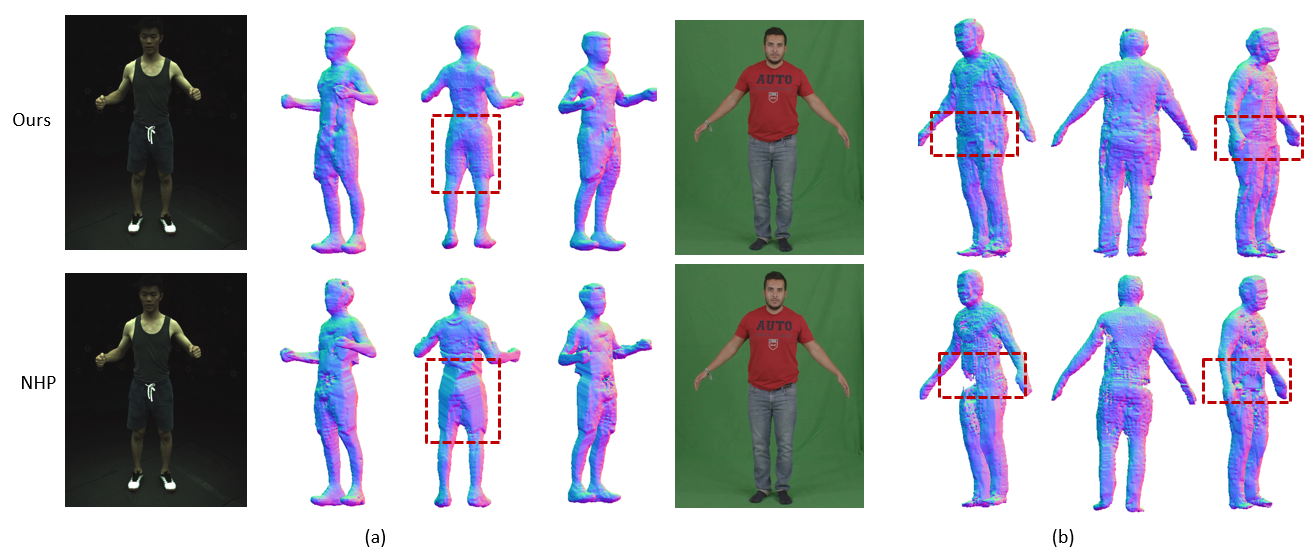}
\end{center}
   \caption{Qualitative comparison for 3D reconstruction on the ZJU-MoCap dataset (a) and People-Snapshot dataset (b).} 
\label{fig:qualitative-results-3d-supp}
\end{figure*}

\paragraph{More qualitative results.}
We compare with NHP \cite{kwon2021neural} for novel view synthesis on the ZJU-Mocap dataset. The results 
for both unseen 
and seen identities are shown in Figure \ref{fig:qualitative-results-unseen-supp} and \ref{fig:qualitative-results-seen-supp} 
, respectively. As marked in red box, we can see that our approach is able to generate more details on the face, arms and legs.  We also show qualitative comparison for 3D reconstruction in Figure \ref{fig:qualitative-results-3d-supp}, where our method reconstructs more realistic shapes as marked in red box.  We also provide more qualitative results for both novel view synthesis and 3D reconstruction of our method in the video.

\paragraph{Inference efficiency}
We introduce the surface-guided sampling strategy to replace the random sampling in our GHuNeRF to improve both accuracy and efficiency. We verify this by comparing with NHP on the People-Snapshot dataset, which consists of 164 testing images. The average rendering quality and per-frame inference time of both methods are shown in Table \ref{inference_time}.
The better rendering quality and faster inference time demonstrate the effectiveness of our approach.
\begin{table}[h!]
\centering
\small
\setlength{\tabcolsep}{10pt}
\begin{tabular*}{0.485\textwidth}{c|c c | c }
\hline
 \multirow{2}{*}{Method} & \multicolumn{2}{c}{Accuracy} & Inference time\\
 & PSNR ($\uparrow$) & SSIM ($\uparrow$) & (s/frame) \\ \hline
NHP & 26.22 & 0.903 & 9.5 \\ 
Ours & 28.37 & 0.927 & 3.5 \\\hline
\end{tabular*}
\caption{Results for both rendering quality and inference time.}
\label{inference_time}
\end{table}

\begin{figure}[h!]
\begin{center}
\includegraphics[width=0.99\linewidth]{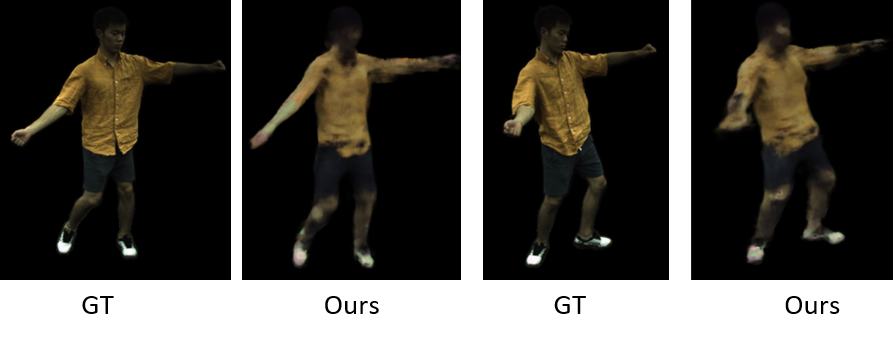}
\end{center} \vspace{-5mm}
   \caption{Examples of failure case.}
\label{fig:qualitative-results-fail-supp}
\end{figure}

\paragraph{Failure cases.}
Our proposed GHuNeRF achieves cross-dataset generalization as demonstrated in Section 4.4 of the main paper. However, the generalization capacity is still limited when the training and testing datasets are significantly different, as mentioned in the limitations. We show some examples in Figure \ref{fig:qualitative-results-fail-supp}, where we test our model trained on the People-Snapshot dataset on the ZJU-MoCap dataset. We can see that our method fails to predict the correct color on the face and legs. The main reason is the significant difference of lighting condition between the People-Snapshot dataset and the ZJU-MoCap dataset. The lighting for the ZJU-MoCap dataset is dark as can be seen from the ground truth images of the results for novel view synthesis, while the lighting for the People-Snapshot dataset is much brighter. Moreover, the yellow 
shirt has never been seen from the People-Snapshot dataset.

{\small
\bibliographystyle{ieee_fullname}
\bibliography{egpaper_final}
}

\end{document}